\PassOptionsToPackage{usenames, svgnames}{xcolor}
\documentclass[final]{article}
\usepackage{iclr2018_conference, times}

\usepackage[T1]{fontenc}
\usepackage[utf8]{inputenc}
\usepackage{microtype}
\usepackage{textcomp}

\usepackage{amsmath, amsthm, amsfonts, amssymb}
\usepackage{thmtools}
\usepackage{mathtools}

\usepackage{graphicx}
\usepackage[hypcap=true]{subcaption}
\usepackage{tcolorbox}
\usepackage{adjustbox}

\usepackage{multirow}
\usepackage{longtable}

\usepackage[inline]{enumitem}
\usepackage{booktabs}
\usepackage[super]{nth}
\usepackage{bm}
\usepackage{pifont}

\usepackage{relsize, scalerel}
\usepackage{twoopt, etoolbox}
\usepackage{setspace}
\usepackage{xspace}
\usepackage{hyphenat}
\usepackage[xspace]{ellipsis}
\usepackage[color=red!55, textsize=footnotesize, textwidth=1.3cm,
            shadow, prependcaption, obeyFinal, colorinlistoftodos]{todonotes}

\usepackage{tikz}
\usetikzlibrary{fit, positioning, arrows.meta, calc}

\usepackage{algorithm}
\usepackage[noend]{algpseudocode}
\usepackage{listings}
\usepackage{varwidth}
\lstalias[]{csharp}[Sharp]{C}
\lstdefinelanguage{dsl}{
    morekeywords = {language,feature,using,semantics,learners,int,string,@input,@start,values,let,in,@ref,@values,Tuple,@extern,@output,namespace,bool,std,@id,grammar},
    otherkeywords = {:=,=>,:,[]},
    sensitive = true,
    morecomment = [l]{//},
    morestring = [b]',
}
\definecolor{linenum-gray}{HTML}{888888}

\algnewcommand\algorithmicbreak{\textbf{break}}
\algnewcommand\Break{\algorithmicbreak\ }
\newcommand{\NoNumber}{\def\alglinenumber##1{}}
\newcommand{\WithNumber}{\def\alglinenumber##1{\sf\scriptsize\color{linenum-gray}##1:\hspace{-11pt}}}
\newcommand{\Functionx}[2]{\NoNumber \Function{#1}{#2} \addtocounter{ALG@line}{-1} \WithNumber}
\let\oldStatex\Statex
\renewcommand{\Statex}{\oldStatex \hspace{\algorithmicindent}}

\usepackage{url}
\usepackage[breaklinks=true, colorlinks=true, citecolor=blue]{hyperref}

\usepackage[capitalize]{cleveref}
\crefname{defn}{definition}{definitions}

\makeatletter
    \pretocmd{\NAT@citexnum}{\@ifnum{\NAT@ctype>\z@}{\let\NAT@hyper@\relax}{}}{}{}
    \DeclareRobustCommand\em
        {\@nomath\em \ifdim \fontdimen\@ne\font >\z@
        \bfseries \else \itshape \fi}
    \renewcommand{\@todonotes@todolistname}{List of TODOs}
\makeatother
\theoremstyle{plain}
\newtheorem{problem}{Problem}
\theoremstyle{definition}

\newtheorem{example}{Example}

\DeclareMathOperator*{\argmax}{\arg\!\max}

\title{Neural\hyp{}Guided Deductive Search for Real\hyp{}Time Program Synthesis from Examples}

\author{%
    Ashwin K. Vijayakumar\thanks{Work done during an internship at Microsoft Research.}\,\,\thanks{Equal contribution.}\,\, \& Dhruv Batra \\
    School of Interactive Computing \\
    Georgia Tech \\
    Atlanta, GA 30308, USA \\
    \texttt{\{ashwinkv,dbatra\}@gatech.edu} \\
    \AND
    Abhishek Mohta\footnotemark[2]\,\, \& Prateek Jain \\
    Microsoft Research India \\
    Bengaluru, Karnataka 560001, India \\
    \texttt{\{t-abmoht,prajain\}@microsoft.com} \\
    \And
    Oleksandr Polozov \& Sumit Gulwani \\
    Microsoft Research Redmond \\
    Redmond, WA 98052, USA \\
    \texttt{\{polozov,sumitg\}@microsoft.com}
}

\iclrfinalcopy

\begin{document}
\newcommand{\PROSE}{\text{PROSE}\xspace}

\newcommand{\dsl}{\ensuremath{\mathcal{L}}\xspace}
\newcommand{\constraint}{\ensuremath{\psi}\xspace}
\newcommand{\state}{\ensuremath{\sigma}\xspace}
\newcommand{\spec}{\ensuremath{\varphi}\xspace}
\newcommand{\Specs}{\ensuremath{\Phi}\xspace}
\newcommand{\rank}{\ensuremath{h}\xspace}
\newcommand{\states}[1][\state]{\ensuremath{\vec{#1}}\xspace}
\newcommand{\Reals}{\ensuremath{\mathbb{R}}}
\newcommand{\tospec}{\ensuremath{\rightsquigarrow}\xspace}
\newcommand{\prog}{\ensuremath{P}\xspace}
\newcommand{\vsa}{\ensuremath{\mathcal{S}}\xspace}
\newcommand{\production}{\ensuremath{\varGamma}\xspace}
\newcommand{\search}{\ensuremath{\mathcal{S}}\xspace}
\newcommand{\learn}{\ensuremath{\mathsf{Learn}}\xspace}
\newcommand{\todoinline}[1]{\vspace*{\topsep}\todo[inline]{\textbf{\textsf{TODO}}: #1}}
\newcommand{\concat}{\mathrel{\circ}}
\newcommand{\palt}{\;|\;}
\newcommand{\pluseq}{\mathrel{{+}{=}}}
\def\etc{\emph{etc}\onedot}
\def\ie{\emph{i.e.}}
\newcommand{\xhdr}[1]{\emph{\textbf{#1}}}

\newcommand{\stringliteral}[1]{\text{\rmfamily\begin{normalsize}``\texttt{#1}''\end{normalsize}}}
\newcommand{\stringliteralwrap}[1]{\begin{normalsize}``\texttt{#1}''\end{normalsize}}
\robustify{\stringliteral}
\robustify{\stringliteralwrap}
\newcommand{\pj}[1]{{\color{red} PJ: #1}}
\newcommand{\ak}[1]{{\color{red} #1}}

\newcommand{\cmark}{\ding{51}}%
\newcommand{\xmark}{\ding{55}}%

\maketitle

\begin{abstract}
Synthesizing user-intended programs from a small number of input-output examples is a challenging problem with several important applications like spreadsheet manipulation, data wrangling and code refactoring.
Existing synthesis systems either completely rely on deductive logic techniques that are extensively hand-engineered or on purely statistical models that need massive amounts of data, and in general fail to provide real-time synthesis on challenging benchmarks.
In this work, we propose \emph{Neural Guided Deductive Search} (NGDS), a hybrid synthesis technique that combines the best of both symbolic logic techniques and statistical models.
Thus, it produces programs that satisfy the provided specifications by construction and generalize well on unseen examples, similar to data-driven systems.
Our technique effectively utilizes the deductive search framework to reduce the learning problem of the neural component to a simple supervised learning setup.
Further, this allows us to both train on sparingly available real-world data and still leverage powerful recurrent neural network encoders.
We demonstrate the effectiveness of our method by evaluating on real-world customer scenarios by synthesizing accurate
programs with up to 12$\times$ speed-up compared to state-of-the-art systems.
\end{abstract}

\vspace{-10pt}
\section{Introduction}
\label{sec:intro}
Automatic synthesis of programs that satisfy a given specification is a classical problem in
AI~\citep{waldinger1969prow}, with extensive literature in both machine learning and programming languages communities.
Recently, this area has gathered widespread interest, mainly spurred by the emergence of a sub-area -- \emph{Programming by
Examples} (PBE) \citep{gulwani_popl11}.
A PBE system synthesizes programs that map a given set of example inputs to their specified example outputs.
Such systems make many tasks accessible to a wider audience as example-based specifications can be easily provided even
by end users without programming skills.
See \Cref{fig:pbe} for an example.
PBE systems are usually evaluated on three key criteria:
\begin{figure}[ht]
    \begin{minipage}{.375\textwidth}
        \begin{tabular}{ll}
            \toprule
            {\bf Input} & {\bf Output}\\ \midrule
            Yann LeCunn & Y LeCunn\\
            Hugo Larochelle & H Larochelle\\
            Tara Sainath & T Sainath\\ \midrule
            {\em Yoshua Bengio} & \multicolumn{1}{c}{?}\\
            \bottomrule
        \end{tabular}
    \end{minipage}
    \begin{minipage}{.624\textwidth}
        \caption{An example input-output spec;
            the goal is to learn a program that maps the given inputs to the corresponding outputs \emph{and}
            generalizes well to new inputs.
            Both programs below satisfy the spec:
            \textbf{(i)} $\mathsf{Concat}$($1^{\text{st}}$ letter of $1^{\text{st}}$ word, $2^{\text{nd}}$ word),
            \textbf{(ii)} $\mathsf{Concat}$($4^{\text{th}}$-last letter of $1^{\text{st}}$ word, $2^{\text{nd}}$ word).
            However, program \textbf{(i)} clearly generalizes better: for instance, its output on ``Yoshua Bengio'' is
            ``Y Bengio'' while program \textbf{(ii)} produces ``s Bengio''.}
        \label{fig:pbe}
    \vspace{-1.2\baselineskip}
	\end{minipage}
\end{figure}
\textbf{(a)} \emph{correctness}: whether the synthesized program satisfies the spec \ie\ the provided example input-output mapping,
\textbf{(b)} \emph{generalization}: whether the program produces the desired outputs
on \emph{unseen} inputs,
and finally, \textbf{(c)} \emph{performance}: synthesis time.

State-of-the-art PBE systems are either \emph{symbolic}, based on enumerative or deductive
search~\citep{gulwani_popl11,polozov_oopsla15} or \emph{statistical}, based on data-driven learning to induce the most
likely program for the spec~\citep{gaunt_arxiv16, balog_iclr16, devlin_icml17}.
Symbolic systems are designed to produce a correct program \emph{by construction} using logical reasoning and
domain-specific knowledge.
They also produce the {\em intended} program with few input-output examples (often just 1).
However, they require significant engineering effort and their underlying search processes struggle with real-time
performance, which is critical for user-facing PBE scenarios.

In contrast, statistical systems do not rely on specialized deductive algorithms, which makes their implementation and
training easier.
However, they lack in two critical aspects.
First, they require a lot of training data and so are often trained using {\em randomly} generated tasks.
As a result, induced programs can be fairly unnatural and fail to generalize to real-world tasks with a small number of
examples.
Second, purely statistical systems like RobustFill~\citep{devlin_icml17} do not \emph{guarantee} that the generated
program satisfies the spec.
Thus, solving the synthesis task requires generating multiple programs with a beam search and post-hoc filtering,
which defeats real-time performance.

\paragraph{Neural-Guided Deductive Search}
Motivated by shortcomings of both the above approaches, we propose \emph{Neural-Guided Deductive Search} (NGDS), a
hybrid synthesis technique that brings together the desirable aspects of both methods.
The symbolic foundation of NGDS is \emph{deductive search}~\citep{polozov_oopsla15} and is parameterized by an underlying
\emph{domain-specific language} (DSL) of target programs.
Synthesis proceeds by recursively applying production rules of the DSL to
decompose the initial synthesis problem into {\em smaller} sub-problems and further applying the same search technique on them.
Our \textbf{key observation I} is that most of the deduced sub-problems do not contribute to the final best program and
therefore \emph{a priori} predicting the usefulness of pursuing a particular sub-problem streamlines the search
process resulting in considerable time savings.
In NGDS, we use a statistical model trained on real-world data to predict a score that corresponds to the likelihood of
finding a \emph{generalizable} program as a result of exploring a sub-problem branch.

Our \textbf{key observation II} is that speeding up deductive search while retaining its correctness or generalization
requires a close integration of
symbolic and statistical approaches via an intelligent controller.
It is based on the ``branch \& bound'' technique from combinatorial optimization~\citep{clausen_branchnbound}.
The overall algorithm integrates (i) deductive search, (ii) a statistical model that predicts,
\emph{a priori}, the generalization score of the best program from a branch, and (iii) a controller that selects
sub-problems for further exploration based on the model's predictions.


Since program synthesis is a sequential process wherein a sequence of decisions (here, selections of DSL rules)
collectively construct the final program, a reinforcement learning setup seems more natural.
However, our \textbf{key observation III} is that deductive search is \emph{Markovian} -- it generates \emph{independent} sub-problems at every level.
In other words, we can reason about a satisfying program for the sub-problem without factoring in the bigger problem from which it was deduced.
This brings three benefits enabling a \emph{supervised learning} formulation:
\textbf{(a)} a dataset of search decisions at every level over a relatively small set of PBE tasks that contains an
exponential amount of information about the DSL promoting generalization,
\textbf{(b)} such search traces can be generated and used for \emph{offline} training,
\textbf{(c)} we can learn separate models for different classes of sub-problems (e.g. DSL levels or rules), with
relatively simpler supervised learning tasks.
\paragraph{Evaluation}
We evaluate NGDS on the string transformation domain, building on top of PROSE, a commercially successful
deductive synthesis framework for PBE~\citep{polozov_oopsla15}.
It represents one of the most widespread and challenging applications of PBE and has shipped in multiple mass-market
tools including Microsoft Excel and Azure ML Workbench.\footnote{\url{https://microsoft.github.io/prose/impact/}}
We train and validate our method on $375$ scenarios obtained from real-world customer tasks
\citep{gulwani_popl11,devlin_icml17}.
Thanks to the Markovian search properties described above, these scenarios generate a dataset of $400,000+$ intermediate
search decisions.
NGDS produces intended programs on $68\%$ of the scenarios despite using only {\em one} input-output example.
In contrast, state-of-the-art neural synthesis techniques \citep{balog_iclr16,devlin_icml17} learn intended
programs from a single example in only $24$-$36\%$ of scenarios taking $\approx 4\times$ more time.
Moreover, NGDS  matches the accuracy of baseline PROSE while providing a speed-up of up to $12\times$ over challenging
tasks.

\paragraph{Contributions}
First, we present a  branch-and-bound optimization based controller that exploits deep neural network based
score predictions to select grammar rules efficiently (\Cref{sec:model:selection}).
Second, we propose a program synthesis algorithm that combines key traits of a symbolic and a statistical approach
to retain desirable properties like
correctness, robust generalization, and real-time performance (\Cref{sec:model:algorithm}).
Third, we evaluate NGDS against state-of-the-art baselines on real customer tasks and show significant gains
(speed-up of up to $12\times$) on several critical cases (\Cref{sec:evaluation}).

\section{Background}
\label{sec:background}
In this section, we provide a brief background on PBE and the PROSE framework, using established
formalism from the programming languages community.

\paragraph{Domain-Specific Language}
A program synthesis problem is defined over a \emph{domain-specific language} (DSL).
A DSL is a restricted programming language that is suitable for expressing tasks in a given domain, but small enough to restrict a search space for program synthesis.
For instance, typical real-life DSLs with applications in textual data transformations~\citep{gulwani_popl11}
often include conditionals, limited forms of loops, and domain-specific operators such as string concatenation, regular
expressions, and date/time formatting.
DSLs for tree transformations such as code refactoring~\citep{rolim2017learning} and data
extraction~\citep{le2014flashextract} include list/data-type processing operators such as $\mathsf{Map}$ and
$\mathsf{Filter}$, as well as domain-specific matching operators.
Formally, a DSL \dsl is specified as a context-free grammar, with each non-terminal symbol $N$ defined by a
set of productions.
The right-hand side of each production is an application of some operator $F(N_1, \dots, N_k)$ to some symbols of~\dsl.
All symbols and operators are strongly typed.
\Cref{fig:background:ffdsl} shows a subset of the Flash Fill DSL that we use as a running example in this paper.

\paragraph{Inductive Program Synthesis}
The task of inductive program synthesis is characterized by a \emph{spec}.
A spec \spec is a set of $m$ input-output \emph{constraints} $ \{ \state_i \tospec \constraint_i \}_{i=1}^m $, where:
\begin{itemize}[topsep=0pt, noitemsep,wide=0pt, leftmargin=\dimexpr\labelwidth + 2\labelsep\relax]
    \item \state, an \emph{input state} is a mapping of free variables of the desired program $P$ to some
        correspondingly typed values.
        At the top level of \dsl, a program (and its expected input state) has only one free variable -- the \emph{input
        variable} of the DSL (e.g., $inputs$ in \Cref{fig:background:ffdsl}).
        Additional local variables are introduced inside \dsl with a \texttt{let} construct.
    \item \constraint is an \emph{output constraint} on the execution result of the desired program $P(\state_i)$.
        At the top level of \dsl, when provided by the user, \constraint is usually the \emph{output example} --
        precisely the expected result of $P(\state_i)$.
        However, other intermediate constraints arise during the synthesis process.
        For instance, \constraint may be a \emph{disjunction} of multiple allowed outputs.
\end{itemize}
The overall goal of program synthesis is thus: {given a spec \spec, find a program $P$ in the underlying DSL~\dsl that
\emph{satisfies} \spec, \emph{i.e.}, its outputs $P(\state_i)$ satisfy all the corresponding constraints
$\constraint_i$.}

\begin{example}
    Consider the task of formatting a phone number, characterized by the spec
    $\spec = \left\{ inputs\colon \left[\stringliteral{(612) 8729128}\right] \right\} \tospec
        \stringliteral{612\textnormal{-}872\textnormal{-}9128}$.
    It has a single input-output example, with an input state \state containing a single variable $inputs$ and its
    value which is a list with a single input string.
    The output constraint is simply the desired program result.

    The program the user is most likely looking for is the one that extracts (a) the part of the input enclosed in the first
    pair of parentheses, (b) the 7$^\text{th}$ to 4$^\text{th}$ characters from the end, and (c) the last 4
    characters, and then concatenates all three parts using hyphens.
    In our DSL, this corresponds to:
    {\small\vspace{-.2\baselineskip}
    \begin{align*}
        \mathsf{Concat}\bigl(&\mathsf{SubStr_0}(\mathsf{RegexPosition}(x, \left\langle\stringliteral{(}, \varepsilon\right\rangle, 0),
            \mathsf{RegexPosition}(x, \left\langle \varepsilon, \stringliteral{)} \right\rangle, 0)), \quad
            \mathsf{ConstStr}(\stringliteral{\textnormal{-}}), \\
        &\mathsf{SubStr_0}(\mathsf{AbsolutePosition}(x, -8), \mathsf{AbsolutePosition}(x, -5)), \quad
            \mathsf{ConstStr}(\stringliteral{\textnormal{-}}), \\
        &\mathsf{SubStr_0}(\mathsf{AbsolutePosition}(x, -5), \mathsf{AbsolutePosition}(x, -1)) \bigr)
    \end{align*}}%
    where $\varepsilon$ is an empty regex, $\mathsf{SubStr_0}(pos_1, pos_2)$ is an abbreviation for ``$\mathtt{let}\ x =
    \mathsf{std.Kth}(inputs,\, 0)$ $\mathtt{in}\ \mathsf{Substring}(x, \langle pos_1, pos_2 \rangle)$'',
    and $\langle \cdot \rangle$ is an abbreviation for $\mathsf{std.Pair}$.

    However, many other programs in the DSL also satisfy \spec.
    For instance, all occurrences of \stringliteral{8} in the output can be produced via a subprogram that simply
    extracts the last character. 
    Such a program overfits to \spec and is bound to fail for other
    inputs where the last character and the $4^{\text{th}}$ one differ.
    \label{ex:background:ff}
\end{example}

As \Cref{ex:background:ff} shows, typical real-life problems are severely underspecified.
A DSL like FlashFill may contain up to $10^{20}$ programs
that satisfy a given spec of 1-3 input-output examples~\citep{polozov_oopsla15}.
Therefore, the main challenge lies in finding a program that not only satisfies the provided input-output examples but
also generalizes to {\em unseen inputs}.
Thus, the synthesis process usually interleaves \emph{search} and \emph{ranking}: the search phase finds a set of
{\em spec-satisfying} programs in the DSL, from which the ranking phase selects top programs ordered using a
domain-specific ranking function $h\colon \dsl \times \vec{\Sigma} \to \Reals$ where $\Sigma$ is the set of all
input states.
The ranking function takes as input a candidate program $P \in \dsl$ and a set of input states $\vec{\state} \in
\vec{\Sigma}$ (usually $\vec{\state} =$ inputs in the given spec + any available unlabeled inputs), and produces a score
for $P$'s
\emph{generalization}.

The implementation of \rank expresses a subtle balance between program generality, complexity, and behavior on available
inputs.
For instance, in FlashFill \rank penalizes overly specific regexes, prefers programs
that produce fewer empty outputs, and prioritizes lower Kolmogorov complexity, among other features.
In modern PBE systems like PROSE, \rank is usually learned in a data-driven manner from customer
tasks~\citep{singh_cav15,ellis2017learning}.
While designing and learning such a ranking is an interesting problem in itself, in this
work we assume a black-box access to \rank.
Finally, the problem of inductive program synthesis can be summarized as follows:
\begin{problem}
    \setstretch{1.10}
    Given a DSL \dsl, a ranking function \rank, a spec $\spec = \{ \state_i \tospec \constraint_i \}_{i=1}^m$,
    optionally a set of unlabeled inputs $\vec{\sigma}_u$, and a target number of programs $K$, let $\vec{\sigma} =
    \vec{\sigma}_u \cup \{ \state_i \}_{i=1}^m$.
    The goal of \emph{inductive program synthesis} is to find a program set $\vsa = \{
    P_1, \dots, P_K \} \subset \dsl$ such that \textbf{(a)} every program in \vsa satisfies \spec,
    and \textbf{(b)} the programs in \vsa generalize best: $\rank(P_i, \vec{\sigma}) \ge \rank(P, \vec{\sigma})$ for any
    other $P \in \dsl$ that satisfies \spec.
    \label{prob:synthesis}
    \vspace*{-0.7\baselineskip}
\end{problem}

\begin{figure}
    \centering
    \begin{lstlisting}[language=dsl,gobble=8,morekeywords={Regex}]
        // Nonterminals
        @start string $transform$ $\coloneqq$ $atom$ | Concat($atom$, $transform$);
        string $atom$ $\coloneqq$ ConstStr($s$)
                         | let string $x$ = std.Kth($inputs$, $k$) in Substring($x$, $pp$);
        Tuple<int, int> $pp$ $\coloneqq$ std.Pair($pos$, $pos$) | RegexOccurrence($x$, $r$, $k$);
        int $pos$ $\coloneqq$ AbsolutePosition($x$, $k$) | RegexPosition($x$, std.Pair($r$, $r$), $k$);
        // Terminals
        @input string[] $inputs$; $\qquad$ string $s$;  $\qquad$   int $k$;  $\qquad$  Regex r;
    \end{lstlisting}
    \vspace{-1.0\baselineskip}
    \caption{%
        A subset of the FlashFill DSL \citep{gulwani_popl11}, used as a running example in this paper.
        Every program takes as input a list of strings $inputs$, and returns an output string, a
        {\em concatenation} of {\em atoms}.
        Each atom is either a constant or a substring of one of the inputs ($x$), extracted using some position
        logic.
        The $\mathsf{RegexOccurrence}$ position logic finds  $k^{\text{th}}$ occurrence of a regex $r$ in $x$
        and returns its boundaries.
        Alternatively, start and end positions can be selected independently either as absolute indices in $x$
        from left or right ($\mathsf{AbsolutePosition}$) or as the $k^{\text{th}}$ occurrence of a pair of
        regexes surrounding the position ($\mathsf{RegexPosition}$).
        See~\citet{gulwani_popl11} for an in-depth DSL description.
    }
    \label{fig:background:ffdsl}
    \vspace{-7pt}
\end{figure}

\paragraph{Search Strategy}
Deductive search strategy for program synthesis, employed by PROSE
explores the grammar of \dsl \emph{top-down} -- iteratively unrolling the productions into partial programs starting from the root symbol.
Following the divide-and-conquer paradigm, at each step it reduces its synthesis problem to smaller
subproblems defined over the parameters of the current production.
Formally, given a spec \spec and a symbol $N$,
PROSE computes the set $\mathsf{Learn}(N, \spec)$ of top programs w.r.t. \rank using two guiding principles:
\begin{enumerate}[noitemsep,topsep=0pt, wide=0pt, leftmargin=\dimexpr\labelwidth + 2\labelsep\relax]
    \item If $N$ is defined through $n$ productions $N \coloneqq F_1(\dots) \palt \dots \palt F_n(\dots)$, PROSE
        finds a \spec-satisfying program set for {\em every} $F_i$, and unites the results, i.e., $\mathsf{Learn}(N, \spec)=\cup_i\ \mathsf{Learn}(F_i(\dots), \spec)$.
    \item For a given production $N \coloneqq F(N_1, \dots, N_k)$, PROSE spawns off $k$ smaller synthesis problems
        $\mathsf{Learn}(N_j, \spec_j),\ 1 \le j \le k$ wherein PROSE deduces necessary and sufficient specs~$\spec_j$ for
        each $N_j$ such that every program of type $F(P_1, \dots, P_k)$, where $P_j \in \mathsf{Learn}(N_j, \spec_j)$, satisfies $\spec$. 
        The deduction logic (called a \emph{witness function}) is domain-specific for each operator $F$.
        PROSE then again recursively solves each subproblem and unites a \emph{cross-product} of the results.
\end{enumerate}
\begin{example}
    Consider a spec $\spec = \{ \stringliteral{Yann} \tospec \stringliteral{Y.L} \}$ on a $transform$ program.
    Via the first production $transform \coloneqq atom$, the only \spec-satisfying program is
    $\mathsf{ConstStr}(\stringliteral{Y.L})$.
    The second production on the same level is $\mathsf{Concat}(atom,\, transform)$.
    A necessary \& sufficient spec on the $atom$ sub-program is that it should produce \emph{some prefix} of the output
    string.
    Thus, the witness function for the $\mathsf{Concat}$ operator produces a \emph{disjunctive spec}
    $\spec_a = \{ \stringliteral{Yann} \tospec \stringliteral{Y} \vee \stringliteral{Y.} \}$.
    Each of these disjuncts, in turn, induces a corresponding necessary and sufficient \emph{suffix} spec on the second
    parameter:
    $\spec_{t1} = \{ \stringliteral{Yann} \tospec \stringliteral{.L} \}$, and
    $\spec_{t2} = \{ \stringliteral{Yann} \tospec \stringliteral{L} \}$, respectively.
    The disjuncts in $\spec_a$ will be recursively satisfied by different program sets:
    \stringliteral{Y.} can only be produced via an $atom$ path with
    a $\mathsf{ConstStr}$ program, whereas \stringliteral{Y} can also be \emph{extracted} from the input using many
    $\mathsf{Substring}$ logics (their generalization capabilities vary).
    \Cref{fig:background:example} shows the resulting search DAG.
    \label{ex:background:search}
\end{example}
\begin{figure}[ht]
    \small
    \begin{tikzpicture}[transform shape, remember picture,
        symbol/.style={draw=SandyBrown, fill=SandyBrown!65, rounded corners=5pt, align=center},
        operator/.style={draw=Blue!50, fill=Blue!15, align=center},
        recurse/.style={-{Triangle[scale=0.5]}, line width=1pt, dotted, draw=Gray}]
        \node[symbol] (t-root) {$transform$ \\ \stringliteral{Y.L}};
        \node[operator, right=12pt of t-root] (c-root) {$\mathsf{Concat}(\dots)$ \\ \stringliteral{Y.L}};
        \node[symbol, below=13pt of t-root] (a-root) {$atom$ \\ \stringliteral{Y.L}};
        \draw[recurse] (t-root) -- (a-root);
        \draw (t-root) -- (c-root);

        \node[symbol, right=72pt of c-root] (a-dis) {$atom$ \\ $\stringliteral{Y} \vee \stringliteral{Y.}$};
        \draw[recurse] (c-root) -- (a-dis);
        \node[symbol, below=13pt of a-dis] (t-1) {$transform$ \\ \stringliteral{L}};
        \node[symbol, below=13pt of t-1] (a-1) {$atom$ \\ \stringliteral{L}};
        \draw[recurse] (t-1) -- (a-1);

        \node[symbol, below=13pt of c-root] (t-2) {$transform$ \\ \stringliteral{.L}};
        \node[symbol, below=13pt of t-2] (a-2) {$atom$ \\ \stringliteral{.L}};
        \draw[recurse] (t-2) -- (a-2);

        \node[operator, right=10pt of t-2] (c-2) {$\mathsf{Concat}(\dots)$ \\ \stringliteral{.L}};
        \draw (t-2) -- (c-2);
        \node[symbol, below=13pt of c-2] (a-dot) {$atom$ \\ \stringliteral{.}};
        \draw[recurse] (c-2) -- (a-dot);

        \draw[recurse] (c-root) -- (t-2);
        \draw[recurse] (c-2) -- (t-1);
        \draw[recurse] ($(c-root)!0.45!(t-2)$) -| (t-1);

        \node[operator, below=13pt of a-root] (const-root) {$\mathsf{ConstStr}(s)$ \\ \stringliteral{Y.L}};
        \node[right=5pt of const-root] (dots-root) {\dots};
        \draw (a-root) -- (const-root);
        \draw ($(a-root)!0.45!(const-root)$) -| (dots-root);

        \node[right=15pt of a-2] (dots-2) {\dots};
        \draw (a-2) -- (dots-2);
        \node[right=15pt of a-dot] (dots-dot) {\dots};
        \draw (a-dot) -- (dots-dot);
        \node[right=15pt of a-1] (dots-1) {\dots};
        \draw (a-1) -- (dots-1);

        \node[operator, right=13pt of a-dis] (const-dis) {$\mathsf{ConstStr}(s)$ \\ $\stringliteral{Y} \vee
            \stringliteral{Y.}$};
        \draw (a-dis) -- (const-dis);
        \node[operator, below=13pt of const-dis] (let-dis) {\texttt{let} $x$ = \dots \\ $\stringliteral{Y} \vee
            \stringliteral{Y.}$};
        \draw ($(a-dis)!0.5!(const-dis)$) |- (let-dis);

        \node[right=8pt of let-dis] (dots-let) {\vdots};
        \node[operator, right=8pt of dots-let] (substring) {$\mathsf{Substring}(\dots)$ \\ \stringliteral{Y}};
        \node[symbol, below=13pt of substring] (pp) {$pp$ \\ $(0, 1)$};
        \draw[recurse] (let-dis) -- (dots-let) -- (substring);
        \draw[recurse] (substring) -- (pp);
        \node[right=5pt of pp] (dots-pp) {\dots};
        \draw[recurse] (pp) -- (dots-pp);
    \end{tikzpicture}
    \caption{A portion of the search DAG from \Cref{ex:background:search}.
        Only the output parts of the respective specs are shown in each node, their common input state is a single
        string \stringliteral{Yann}.
        Dashed arrows show recursive $\mathsf{Learn}$ calls on a corresponding DSL symbol.
    }
    \label{fig:background:example}
\end{figure}

Notice that the above mentioned principles create {\em  logical non-determinism} due to which we might need to explore multiple alternatives in a search tree.
As such non-determinism arises at every level of the DSL with potentially any operator, the search tree (and the
resulting search process) is exponential in size.
While all the branches of the tree by construction produce programs that satisfy the given spec, most
of the branches do not contribute to the overall top-ranked \emph{generalizable} program.
During deductive search, PROSE has limited information about the programs potentially produced
from each branch, and cannot estimate their quality, thus exploring the entire tree unnecessarily.
Our main contribution is a \emph{neural\hyp{}guided search algorithm} that predicts the best program scores
from each branch, and allows PROSE to omit branches that are unlikely to produce the desired program \emph{a priori}.

\section{Synthesis Algorithm}
\label{sec:model}

Consider an arbitrary branching moment in the top-down search strategy of PROSE.
For example, let $N$ be a nonterminal symbol in \dsl, defined through a set of productions $N \coloneqq F_1(\dots) \palt
\dots \palt F_n(\dots)$, and let \spec be a spec on $N$, constructed earlier during the recursive descent over \dsl.
A conservative way to select the top $k$ programs rooted at $N$ (as defined by the ranking function \rank), i.e., to compute $\learn(N, \spec)$, is to
learn the top $k$ programs of kind $F_i(\dots)$ for all $i \in [k]$ and then
select the top $k$ programs overall from the union of program sets learned for each production.
Naturally, exploring all the branches for each nonterminal in the search tree is computationally expensive.

In this work, we propose a data-driven method to select an appropriate production rule $N\coloneqq F_i(N_1, \dots, N_k)$
that would most likely lead to a top-ranked program.
To this end, we use the current spec \spec to determine the ``optimal'' rule.
Now, it might seem unintuitive that even without exploring a production rule and finding the best program in the
corresponding program set, we can \emph{a priori} determine optimality of that rule.
However, we argue that by understanding \spec and its relationship with the ranking function \rank, we can
\emph{predict} the intended branch in many real-life scenarios.

\begin{example}
Consider a spec $\spec=\{\stringliteral{alice} \tospec \stringliteral{alice@iclr.org},\ \stringliteral{bob} \tospec
\stringliteral{bob@iclr.org}\}$.
While learning a program in \dsl given by \Cref{fig:background:ffdsl} that satisfies~\spec, it is clear right
at the beginning of the search procedure that the rule $transform \coloneqq atom$ does not apply.
This is because any programs derived from $transform \coloneqq atom$ can either extract a substring from the input or
return a constant string, both of which fail to produce the desired output.
Hence, we should only consider $transform \coloneqq \mathsf{Concat}(\dots)$, thus significantly reducing the search
space.

Similarly, consider another spec $\spec=\{\stringliteral{alice smith}\tospec \stringliteral{alice},\ \stringliteral{bob
jones}\tospec \stringliteral{bob}\}$.
In this case, the output appears to be a substring of input, thus selecting $transform \coloneqq atom$ at the
beginning of the search procedure is a better option than $transform \coloneqq \mathsf{Concat}(\dots)$.

However, many such decisions are more subtle and depend on the ranking function \rank itself.
For example, consider a spec $\spec=\{\stringliteral{alice liddell}\tospec \stringliteral{al},
\ \stringliteral{bob ong}\tospec \stringliteral{bo}\}$.
Now, both $transform \coloneqq atom$ and $transform \coloneqq \mathsf{Concat}(\dots)$ may lead to viable programs
because the output can be constructed using the first two letters of the input (i.e. a substring atom) or by
\emph{concatenating} the first letters of each word.
Hence, the branch that produces the best program is ultimately determined by the ranking function \rank since both
branches generate valid programs.
\label{ex:model:motivation}
\end{example}

\Cref{ex:model:motivation} shows that to design a data-driven search strategy for branch selection, we
need to learn the subtle relationship between \spec, \rank, and the candidate branch.
Below, we provide one such model.

\subsection{Predicting the Generalization Score}\vspace*{-5pt}
\label{sec:model:score}
As mentioned above, our goal is to predict one or more production rules that for a given spec \spec will lead to a
top-ranked program (as ranked \emph{a posteriori} by \rank).
Formally, given black-box access to $\rank$, we want to learn a function $f$ such that,
\[
    f(\production, \spec)\approx \max_{P \,\in\, \vsa(\production,\, \spec)} \rank(P, \spec),
\]
where $\production$ is a production rule in \dsl, and $\vsa(\production, \spec)$ is a \emph{program set} of all
DSL programs derived from the rule $\production$ that satisfy \spec.
In other words, we want to predict the score of the top-ranked \emph{\spec-satisfying} program that is synthesized by  unrolling the rule $\production$.  We assume that the symbolic search of PROSE handles the construction of $\vsa(\production, \spec)$ and ensures
that programs in it satisfy \spec by construction.
The goal of $f$ is to optimize the score of a program derived from \production \emph{assuming} this program is valid.
If no program derived from \production can satisfy \spec, $f$ should return $-\infty$. Note that, drawing upon observations mentioned in Section~\ref{sec:intro}, we have cast the production selection problem as a {\em supervised learning} problem, thus simplifying the learning task  as opposed to end-to-end reinforcement learning solution. 

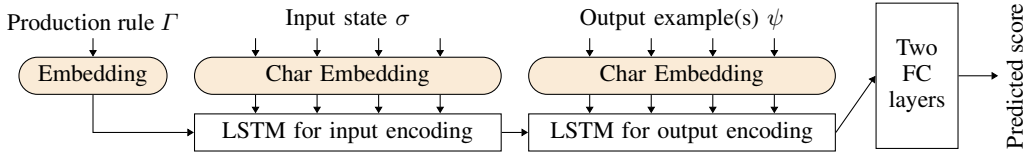
\begin{figure}[t!]
    \centering
    \small
    \hspace{-1em}
\begin{tikzpicture}[transform shape, remember picture,
                    connection/.style={-{Triangle[scale=0.7]}}]
    \node[draw, align=center, text width=110pt] (input-enc) {LSTM for input encoding};
    \node[draw, right=10pt of input-enc, align=center, text width=110pt] (output-enc) {LSTM for output encoding};
    \draw[connection] (input-enc) -- (output-enc);

    \node[draw, above=7pt of input-enc, rounded corners=7pt, text width=110pt, align=center, fill=AntiqueWhite]
        (emb-input) {Char Embedding};

    \node[above=7pt of emb-input, align=center, text width=110pt] (input) {Input state \state};
    \foreach \i in {0.2, 0.4, 0.6, 0.8} {
        \draw[connection] ($(input.south west)!\i!(input.south east)$) -- ($(emb-input.north west)!\i!(emb-input.north east)$);
        \draw[connection] ($(emb-input.south west)!\i!(emb-input.south east)$) -- ($(input-enc.north west)!\i!(input-enc.north east)$);
    }

    \node[draw, above=7pt of output-enc, rounded corners=7pt, text width=110pt, align=center, fill=AntiqueWhite]
        (emb-output) {Char Embedding};
    \node[above=7pt of emb-output, align=center, text width=110pt] (output) {Output example(s) \constraint};
    \foreach \i in {0.2, 0.4, 0.6, 0.8} {
        \draw[connection] ($(output.south west)!\i!(output.south east)$) -- ($(emb-output.north west)!\i!(emb-output.north east)$);
        \draw[connection] ($(emb-output.south west)!\i!(emb-output.south east)$) -- ($(output-enc.north west)!\i!(output-enc.north east)$);
    }

    \node[draw, left=10pt of emb-input, rounded corners=7pt, align=center, text width=50pt, fill=AntiqueWhite] (emb-rule) {Embedding};
    \draw[connection] (emb-rule) |- (input-enc);
    \node[above=7pt of emb-rule] (rule) {Production rule \production};
    \draw[connection] (rule) -- (emb-rule);

    \node[draw, right=15pt of emb-output, minimum height=55pt, text width=25pt, align=center] (fc) {Two FC layers};
    \draw[connection] (output-enc.east) -- (fc.west);
    \node[right=15pt of fc, anchor=north, rotate=90] (score) {Predicted score};
    \draw[connection] (fc) -- (score);
\end{tikzpicture}
    \vspace{-0.5\baselineskip}
    \caption{LSTM-based model for predicting the score of a candidate production for a given spec \spec.}
	\label{fig:model:lstm}
    \vspace{-0.4\baselineskip}
\end{figure}

We have evaluated two models for learning $f$.
The loss function for the prediction is given by:
\[
    L(f; \production, \spec)=\bigl(f(\production, \spec)-\max_{P \,\in\, \vsa(\production,\, \spec)} \rank(P, \spec)\bigr)^2.
\]
\Cref{fig:model:lstm} shows a common structure of both models we have evaluated.
Both are based on a standard multi-layer LSTM architecture \citep{hochreiter97} and involve \textbf{(a)} embedding the
given spec \spec, \textbf{(b)} encoding the given production rule \production, and \textbf{(c)} a feed-forward network
to output a score $f(\production, \spec)$.
One model attends over input when it encodes the output, whereas another does not.

\subsection{Controller for Branch Selection}
\label{sec:model:selection}
A score model $f$ alone is insufficient to perfectly predict the branches that should be explored at every level. 
Consider again a branching decision moment $N \coloneqq F_1(\dots) \palt \dots \palt F_n(\dots)$ in a search process for
top $k$ programs satisfying a spec \spec.
One na\"ive approach to using the predictions of $f$ is to always follow the highest-scored production rule
$\argmax_i f(F_i, \spec)$.
However, this means that \emph{any single incorrect decision on the path from the DSL root to the desired program will
eliminate that program from the learned program set}.
If our search algorithm fails to produce the desired program by committing to a suboptimal branch anytime  during the
search process, then the user may never discover that such a program exists unless they supply additional input-output example. 

Thus, a branch selection strategy based on the predictions of $f$ must balance a trade-off of \emph{performance} and
\emph{generalization}.
Selecting too few branches (a single best branch in the extreme case) risks committing to an incorrect path early in the
search process and producing a suboptimal program or no program at all.
Selecting too many branches (all $n$ branches in the extreme case) is no different from baseline PROSE and fails to
exploit the predictions of $f$ to improve its performance.

Formally, a \emph{controller} for branch selection at a symbol $N \coloneqq F_1(\dots) \palt \dots \palt F_n(\dots)$
targeting $k$ best programs must \textbf{(a)} predict the expected score of the best program from each program set:
$
    s_i = f(F_i, \spec)\ \ \forall\, 1\leq i\leq n,
$
and \textbf{(b)} use the predicted scores $s_i$ to narrow down the set of productions $F_1, \dots, F_n$ to explore and
to obtain the overall result by selecting a subset of generated programs.
In this work, we propose and evaluate two controllers.
Their pseudocode is shown in \Cref{fig:model:strategies}.

\textbf{Threshold-based:}
Fix a \emph{score threshold} $\theta$, and explore those branches whose predicted score differs by at most
$\theta$ from the maximum predicted score.
This is a simple extension of the na\"ive ``$\argmax$'' controller discussed earlier that also explores any
branches that are predicted ``approximately as good as the best one''.
When $\theta = 0$, it reduces to the ``$\argmax$'' one.

\textbf{Branch \& Bound:}
This controller is based on the ``branch \& bound'' technique in combinatorial
optimization~\citep{clausen_branchnbound}.
Assume the branches $F_i$ are ordered in the descending order of their respective predicted scores $s_i$.
After recursive learning produces its program set $\vsa_i$, the controller proceeds to the next branch
only if $s_{i+1}$ \emph{exceeds the score of the worst program in $\vsa_i$}.
Moreover, it reduces the target number of programs to be learned, using $s_{i+1}$ as a lower bound
on the scores of the programs in $\vsa_i$.
That is, rather than relying blindly on the predicted scores, the controller guides the remaining search process by
accounting for the actual synthesized programs as well. 

\begin{figure}[t]
    \small
    \hspace{-2em}
    \begin{minipage}[t]{0.50\linewidth}
        \begin{algorithmic}[1]
            \Functionx{ThresholdBased}{$\spec, \rank, k, s_1, \text{\dots}, s_n$}
                \State Result set $\vsa^* \gets []$
                \State $i^* \gets \argmax_i s_i$
                \ForAll{$1 \le i \le n$}
                    \If{$\left| s_i - s_{i^*} \right| \le \theta$}
                        \Statex \hspace{2.5em} \Comment{Recursive search}
                        \State $\vsa^* \pluseq \Call{Learn}{F_i, \spec, k}$
                    \EndIf
                \EndFor
                \State \Return the top $k$ programs of $\vsa$ w.r.t. \rank
            \EndFunction
        \end{algorithmic}
    \end{minipage}
    \begin{minipage}[t]{0.63\linewidth}
        \begin{algorithmic}[1]
            \Functionx{BnBBased}{$\spec, \rank, k, s_1, \text{\dots}, s_n$}
                \State Result set $\vsa^* \gets []$; \quad Program target $k' \gets k$
                \State Reorder $F_i$ in the descending order of $s_i$
                \ForAll{$1 \le i \le n$}
                    \State $\vsa_i \gets \Call{Learn}{F_i, \spec, k'}$ \Comment{Recursive search}
                    \State $j \gets \Call{BinarySearch}{s_{i+1}, \mathsf{Map}(h,\vsa_i)}$
                    \State $\vsa^* =\vsa_i^* \cup \vsa_i[0..j]$;\quad $k' \gets k' - j$
                    \If{$k' \le 0$}
                        \Break
                    \EndIf
                \EndFor
                \State \Return $\vsa^*$
            \EndFunction
        \end{algorithmic}
    \end{minipage}
    \caption{
        The controllers for guiding the search process to construct a \emph{most generalizable} \spec-satisfying program
        set \vsa of size $k$ given the $f$-predicted best scores $s_1, \text{\dots}, s_n$ of the productions $F_1,
        \text{\dots}, F_n$.
    }
    \label{fig:model:strategies}
    \vspace{-0.5\baselineskip}
\end{figure}

\begin{figure}[t]
    \small
    \begin{algorithmic}[1]
        \oldStatex \textbf{Given: }
        \begin{varwidth}[t]{0.9\linewidth}
            DSL \dsl, ranking function \rank, controller $\mathcal{C}$ from
            \Cref{fig:model:strategies} (\textsc{ThresholdBased} or \textsc{BnBBased}),
            symbolic search algorithm \Call{Learn}{Production rule \production, spec \spec, target $k$} as in
            PROSE~\citep[Figure 7]{polozov_oopsla15} with all recursive calls to \textsc{Learn} replaced with
            \textsc{LearnNGDS}
        \end{varwidth}
        \vspace{2pt}
        \Functionx{LearnNGDS}{Symbol $N\coloneqq F_1(\dots) \palt \dots \palt F_n(\dots)$, spec \spec, target number of programs $k$}
            \If{$n = 1$}
                \Return \Call{Learn}{$F_1, \spec, k$}
            \EndIf
            \State Pick a score model $f$ based on $\mathsf{depth}(N, \dsl)$
            \State $s_1, \dots, s_n \gets f(F_1, \spec), \dots, f(F_n, \spec)$
            \State \Return $\mathcal{C}(\spec, h, k, s_1, \dots, s_n)$
        \EndFunction
    \end{algorithmic}
    \caption{Neural-guided deductive search over \dsl, parameterized with a branch selection controller $\mathcal{C}$.}
    \label{fig:model:ngds}
    \vspace{-1.0\baselineskip}
\end{figure}

\subsection{Neural-Guided Deductive Search}
\label{sec:model:algorithm}
We now combine the above components to present our unified algorithm for program synthesis.
It builds upon the \emph{deductive search} of the PROSE system, which uses symbolic PL insights in the form of
\emph{witness functions} to construct and narrow down the search space, and a \emph{ranking function} \rank to pick the
most generalizable program from the found set of spec-satisfying ones. 
However, it significantly speeds up the search process by guiding it \emph{a priori} at each branching
decision using the learned score model~$f$ and a branch selection controller, outlined
in \Cref{sec:model:score,sec:model:selection}.
The resulting \emph{neural-guided deductive search} (NGDS) keeps the symbolic insights that construct the search tree
ensuring correctness of the found programs, but explores only those branches of this tree that are likely to
produce the user-intended generalizable program, thus eliminating unproductive search time.

A key idea in NGDS is that the score prediction model $f$ does not have to be the same for all decisions in the
search process.
It is possible to train separate models for different DSL levels, symbols, or even productions.
This allows the model to use different features of the input-output spec for evaluating the fitness of different
productions, and also leads to much simpler supervised learning problems.

\Cref{fig:model:ngds} shows the pseudocode of NGDS.
It builds upon the deductive search of PROSE, but augments every branching decision on a symbol with some
branch selection controller from \Cref{sec:model:selection}.
We present a comprehensive evaluation of different strategies in \Cref{sec:evaluation}.

\section{Evaluation}
\label{sec:evaluation}

\begin{table}[t]
    \centering
    \small
    \begin{tabular}{l r rrr rrr r}
        \toprule
        \textbf{Metric} &
        \textbf{PROSE} & \textbf{DC}$_1$ & \textbf{DC}$_2$ & \textbf{DC}$_3$ &
        \textbf{RF}$_1$ & \textbf{RF}$_2$ & \textbf{RF}$_3$ & \textbf{NGDS} \\
        \midrule
        \textbf{Accuracy (\% of 73)} &
        67.12 &
        35.81 & 47.38 & 62.92 &
        24.53 & 39.72 & 56.41 &
        \textbf{68.49} \\
        \textbf{Speed-up ($\times$ PROSE)} &
        1.00 &
        \textbf{1.82} & 1.53 & 1.42 &
        0.25 & 0.27 & 0.30 &
        1.67 \\
        \bottomrule
    \end{tabular}
    \caption{
        Accuracy and average speed-up of NGDS vs. baseline methods. Accuracies are computed on a test set of $73$ tasks.
        {\em Speed-up} of a method is the geometric mean of its per-task speed-up (ratio of synthesis time of PROSE and
        of the method) when restricted to a subset of tasks with PROSE's synthesis time is $\ge 0.5$ sec.
    }
    \label{tab:results}
    \vspace{-\baselineskip}
\end{table}

In this section, we evaluate our NGDS algorithm over the string manipulation domain with a DSL given by
\Cref{fig:background:ffdsl}; see \Cref{fig:pbe} for an example task.
We evaluate NGDS, its ablations, and baseline techniques on two key metrics:
(a) generalization accuracy on unseen inputs, (b) synthesis time.


\textbf{Dataset.}
We use a dataset of $375$ \emph{tasks} collected from real-world customer string manipulation problems, split into
$65\%$ training, $15\%$ validation, and $20\%$ test data.
Some of the common applications found in our dataset include date/time formatting, manipulating addresses,
modifying names, automatically generating email IDs, etc.
Each task contains about $10$ inputs, of which {\em only one} is provided as the spec to the synthesis system, mimicking industrial applications.
The remaining {\em unseen} examples are used to evaluate generalization performance of the synthesized programs.
After running synthesis of top-1 programs with PROSE on all training tasks, we have collected a dataset of
$\approx$ 400,000 intermediate search decisions, \emph{i.e.}
triples $\langle \text{production } \production, \text{ spec } \spec, \text{ \emph{a posteriori} best
score } h(P,\spec) \rangle$.

\textbf{Baselines.} We compare our method against two state-of-the-art neural synthesis algorithms: RobustFill
\citep{devlin_icml17} and DeepCoder \citep{balog_iclr16}.
For RobustFill, we use the best-performing \emph{Attention-C} model and
use their recommended DP-Beam Search with a beam size of 100 as it seems to perform the best; Table~\ref{table:robustfill} in Appendix~\ref{app:rfill} presents results with different beam sizes. As in the original work, we select the top-1 program ranked according to the generated $\log$-likelihood. 
DeepCoder is a generic framework that allows their neural predictions to be combined with any program synthesis method.
So, for fair comparison, we combine DeepCoder's predictions with PROSE.
We train DeepCoder model to predict a distribution over \dsl's operators and as proposed, use it to guide PROSE synthesis. 
Since both RobustFill and DeepCoder are trained on randomly sampled programs and are not optimized for generalization in the real-world, we include their variants trained with
2 or 3 examples (denoted RF$_m$ and DC$_m$) for fairness, although $m=1$ example is the most important scenario in
real-life industrial usage.


\textbf{Ablations.}
As mentioned in \cref{sec:model}, our novel usage of score predictors to guide the search enables us to have multiple
prediction models and controllers at various stages of the synthesis process.
Here we investigate ablations of our approach with models that specialize in predictions for individual levels
in the search process.
The model $T_1$ is trained for symbol $transform$ (\Cref{fig:background:ffdsl}) when expanded in the first level.
Similarly, $PP$, $POS$ refer to models trained for the $pp$ and $pos$  symbol, respectively.
Finally, we train all our LSTM-based models with CNTK~\citep{cntk} using Adam~\citep{kingma2014adam} with a learning
rate of $10^{-2}$ and a batch size of 32, using early stopping on the validation loss to select the best performing
model (thus, 100-600 epochs).

\def\bbthr{BB$_{0.2}$\xspace}
We also evaluate three controllers: threshold-based (Thr) and branch-and-bound (BB) controllers given in
\Cref{fig:model:strategies}, and a combination of them~-- branch-and-bound with a $0.2$ threshold predecessor (\bbthr).
In \Cref{tab:results,tab:speed_benchmarks} we denote different model combinations as NGDS($f$, $\mathcal{C}$) where $f$
is a symbol-based model and $\mathcal{C}$ is a controller.
The final algorithm selection depends on its accuracy-performance trade-off.
In \Cref{tab:results}, we use NGDS($T_1 + POS$, BB), the best performing algorithm on the test set, although
NGDS($T_1$, BB) performs slightly better on the validation set.

\textbf{Evaluation Metrics.}
{\em Generalization accuracy} is the percentage of test tasks for which the generated program satisfies \emph{all}
unseen inputs in the task.
{\em Synthesis time} is measured as the wall-clock time taken by a synthesis method to find the correct program, median
over 5 runs.
We run all the methods on the same machine with 2.3 GHz Intel Xeon processor, 64GB of RAM, and Windows Server 2016.


\begin{table}[t]
    \small
    \centering
    \begin{tabular}{l rr rr r}
        \toprule
        \multirow{2}{*}{\textbf{Method}} & \multicolumn{2}{c}{\textbf{Validation}} & \multicolumn{2}{c}{\textbf{Test}}
                                         & \multirow{2}{*}{\textbf{\% of branches}} \\
        \cmidrule(lr){2-3} \cmidrule(lr){4-5}
        & \textbf{Accuracy} & \textbf{Speed-up} & \textbf{Accuracy} & \textbf{Speed-up} \\
        \midrule
        PROSE & 70.21 & 1 & 67.12 & 1 & 100.00 \\
        NGDS($T_1$, Thr) & 59.57 & 1.15 & 67.12 & 1.27 & 62.72 \\
        NGDS($T_1$, BB) & 63.83	& 1.58	& 68.49	& 1.22	& 51.78 \\
        NGDS($T_1$, \bbthr) & 61.70 & 1.03    & 67.12   & 1.22    & 63.16 \\
        NGDS($T_1 + PP$, Thr) & 59.57	& 0.76	& 67.12	& 0.97	& 56.41 \\
        NGDS($T_1 + PP$, BB) & 61.70	& 1.05	& 72.60	& 0.89	& 50.22 \\
        NGDS($T_1 + PP$, \bbthr) & 61.70	& 0.72	& 67.12	& 0.86	& 56.43 \\
        NGDS($T_1 + POS$, Thr) & 61.70	& 1.19	& 67.12	& 1.93	& 55.63 \\
        NGDS($T_1 + POS$, BB) & 63.83	& 1.13	& 68.49	& 1.67	& 50.44 \\
        NGDS($T_1 + POS$, \bbthr) & 63.83	& 1.19	& 67.12	& 1.73	& 55.73 \\
        \bottomrule
    \end{tabular}
    \caption{Accuracies, mean speed-ups, and \% of branches taken for different ablations of NGDS.}
    \label{tab:speed_benchmarks}
    \vspace{-1.0\baselineskip}
\end{table}

\textbf{Results.}
\Cref{tab:results} presents generalization accuracy as well as synthesis time speed-up of various methods w.r.t. PROSE.
As we strive to provide real-time synthesis, we only compare the times for tasks which require PROSE more than $0.5$ sec.
Note that, with one example, NGDS and PROSE are significantly more accurate than RobustFill and DeepCoder.
This is natural as those methods are not trained to optimize generalization, but it also highlights
advantage of a close integration with a symbolic system (PROSE) that incorporates deep domain knowledge.
Moreover, on an average, our method saves more than $50\%$ of synthesis time over PROSE.
While DeepCoder with one example speeds up the synthesis even more, it does so at the expense of accuracy, eliminating
branches with \emph{correct} programs in $65\%$ of tasks.

\Cref{tab:speed_benchmarks} presents speed-up obtained by variations of our models and controllers.
In addition to generalization accuracy and synthesis speed-up, we also show a fraction of branches that were selected
for exploration by the controller.
Our method obtains impressive speed-up of $>1.5\times$ in $22$ cases.
One such test case where we obtain $12\times$ speedup is a simple extraction case which is
fairly common in Web mining: $\{\stringliteral{alpha,beta,charlie,delta}\tospec \stringliteral{alpha}\}$.
For such cases, our model
determine $transform:=atom$ to be the correct branch (that leads to the final $\mathsf{Substring}$ based program) and
hence saves time required to explore the entire $\mathsf{Concat}$ operator which is expensive.
Another interesting test
case where we observe $2.7\times$ speed-up is: $\{\stringliteral{457 124th St S, Seattle, WA 98111}\tospec
\stringliteral{Seattle-WA} \}$.
This test case involves learning a $\mathsf{Concat}$ operator initially followed by
$\mathsf{Substring}$ and $\mathsf{RegexPosition}$ operator.
\Cref{sec:benchmark-table} includes a comprehensive table of NGDS performance on all the validation and test tasks.

All the models in \Cref{tab:speed_benchmarks} run \emph{without} attention.
As measured by \emph{score flip accuracies} (\ie\ percentage of correct orderings of branch scores on the same level),
attention-based models perform best, achieving $99.57/90.4/96.4\%$ accuracy on train/validation/test, respectively (as
compared to $96.09/91.24/91.12\%$ for non-attention models).
However, an attention-based model is significantly more computationally expensive at prediction time.
Evaluating it dominates the synthesis time and eliminates any potential speed-ups.
Thus, we decided to forgo attention in initial NGDS and investigate model compression/binarization in future work.

\textbf{Error Analysis.}
As \cref{sec:benchmark-table} shows, NGDS is slower than PROSE on some tasks.
This occurs when the predictions do not satisfy the constraints of the controller \emph{i.e.} all the predicted
scores are within the threshold or they violate the actual scores during B\&B exploration.
This leads to NGDS evaluating the LSTM for branches that were previously pruned.
This is especially harmful when branches pruned out at the very beginning of the search need to be
reconsidered -- as it could lead to evaluating the neural network many times.
While a single evaluation of the network is quick, a search tree involves many evaluations,
and when performance of PROSE is already $< 1$~s, this results in considerable \emph{relative} slowdown.
We provide two examples to illustrate both the failure modes:

\textbf{(a)}~\stringliteralwrap{41.7114830017,-91.41233825683,41.60762786865,-91.63739013671}  \tospec
\stringliteral{41.7114830017}.
The intended program is a simple substring extraction.
However, at depth 1, the predicted score of $\mathsf{Concat}$ is much higher than the predicted score of
$\mathsf{Atom}$, and thus NGDS explores only the $\mathsf{Concat}$ branch.
The found $\mathsf{Concat}$ program is incorrect because it uses absolute position indexes and does not generalize to
other similar extraction tasks.
We found this scenario common with punctuation in the output string, which the model considers a
strong signal for $\mathsf{Concat}$.

\textbf{(b)} \stringliteralwrap{type size =  36: Bartok.Analysis.CallGraphNode type size =  32:
Bartok.Analysis.CallGraphNode CallGraphNode} \tospec \stringliteral{36->32}.
In this case, NGDS correctly explores only the $\mathsf{Concat}$ branch, but the slowdown happens at the $pos$ symbol.
There are many different logics to extract the \stringliteral{36} and \stringliteral{32} substrings.
NGDS explores the $\mathsf{RelativePosition}$ branch first, but the score of the resulting program is less then the
prediction for $\mathsf{RegexPositionRelative}$.
Thus, the B\&B controller explores both branches anyway, which leads to a relative slowdown caused by the network
evaluation time.

\section{Related Work}
\label{sec:related}

{\bf Neural Program Induction} systems synthesize a program by training a {\em new} neural network model to map the example inputs to example outputs \citep{graves_arxiv14,reed_arxiv15, zaremba_icml16}. Examples include Neural Turing Machines~\citep{graves_arxiv14} that can learn simple programs like copying/sorting, work of \cite{kaiser_arxiv15} that can perform more complex computations like binary multiplications, and more recent work of \cite{cai_iclr17} that can incorporate recursions.
%
While we are interested in ultimately producing the right output, all these models need to be re-trained
for a given problem type, thus making them unsuitable for real-life synthesis of \emph{different} programs with
\emph{few} examples.

{\bf Neural Program Synthesis} systems synthesize a program in a given \dsl with a pre-learned neural network. Seminal works of \cite{riedel_arxiv16} and \cite{gaunt_arxiv16}
proposed first producing a  high-level sketch of the program using procedural knowledge, and then synthesizing the program by combining the sketch with  a neural or enumerative synthesis engine. In contrast, R3NN~\citep{parisotto_iclr16} and RobustFill~\citep{devlin_icml17} systems synthesize the program end-to-end using a neural network; \cite{devlin_icml17} show that RobustFill in fact outperforms R3NN.
However, RobustFill does not guarantee generation of {\em spec-satisfying} programs and often requires
more than one example to find the intended program. In fact, our empirical evaluation (\Cref{sec:evaluation}) shows that our hybrid synthesis approach significantly outperforms the purely statistical approach of RobustFill.

DeepCoder \citep{balog_iclr16} is also a hybrid synthesis system that guides enumerative program synthesis
by prioritizing DSL operators according to a spec-driven likelihood distribution on the same. However, NGDS differs from DeepCoder in two important ways:
(a) it guides the search process \emph{at each recursive level} in a top-down {\em goal-oriented} enumeration and thus reshapes the search tree,
(b)~it is trained on real-world data instead of random programs, thus achieving better generalization.

{\bf Symbolic Program Synthesis} has been studied extensively in the PL community~\citep{GulwaniPS17,sygus}, dating back
as far as 1960s~\citep{waldinger1969prow}.
Most approaches employ either bottom-up enumerative search~\citep{transit}, constraint
solving~\citep{torlak2013growing}, or inductive logic programming~\citep{lin2014bias},
and thus scale poorly to real-world industrial applications (e.g. data wrangling applications).
In this work, we build upon deductive search, first studied for synthesis by~\cite{manna71}, and primarily used
for program synthesis from formal logical specifications~\citep{spiral,chaudhari2015combining}.
\cite{gulwani_popl11} and later~\cite{polozov_oopsla15} used it to build PROSE, a commercially successful
domain-agnostic system for PBE.
While its deductive search guarantees program correctness and also good generalization via an accurate ranking function,
it still takes several seconds on complex tasks.
Thus, speeding up deductive search requires considerable engineering to develop manual heuristics.
NGDS instead integrates neural-driven predictions at each level of deductive search to alleviate this drawback.
Work of \cite{irving_arxiv16} represents the closest work with a similar technique but their work is applied to an
automated theorem prover, and hence need not care about generalization.
In contrast, NGDS guides the search toward generalizable programs while relying on the underlying symbolic engine to
generate correct programs.

\section{Conclusion}
\label{sec:conclusion}
We studied the problem of real-time program synthesis with a small number of input-output examples. For this problem, we proposed a neural-guided system that builds upon PROSE, a state-of-the-art symbolic logic based system. Our system avoids top-down {\em enumerative} grammar exploration required by PROSE thus providing impressive synthesis performance while still retaining key advantages of a deductive system. That is, compared to existing neural synthesis techniques, our system enjoys following advantages: a) {\em correctness}: programs generated by our system are guaranteed to satisfy the given input-output specification, b) {\em generalization}: our system learns the user-intended program with just one input-output example in around 60\% test cases while existing neural systems learn such a program in only 16\% test cases, c) {\em synthesis time}: our system can solve most of the test cases in less than 0.1 sec and provide impressive performance gains over both neural as well symbolic systems. 

The key take-home message of this work is that a deep integration of a symbolic deductive inference based system with statistical techniques leads to best of both the worlds where we can avoid extensive engineering effort required by symbolic systems without compromising the quality of generated programs, and at the same time provide significant performance (when measured as synthesis time) gains. For future work, exploring better learning models for production rule selection and applying our technique to diverse and  more powerful grammars should be important research directions.

\bibliography{ml2search}

\begin{thebibliography}{31}
\providecommand{\natexlab}[1]{#1}
\providecommand{\url}[1]{\texttt{#1}}
\expandafter\ifx\csname urlstyle\endcsname\relax
  \providecommand{\doi}[1]{doi: #1}\else
  \providecommand{\doi}{doi: \begingroup \urlstyle{rm}\Url}\fi

\bibitem[Alur et~al.(2013)Alur, Bod{\'{\i}}k, Juniwal, Martin, Raghothaman,
  Seshia, Singh, Solar{-}Lezama, Torlak, and Udupa]{sygus}
Rajeev Alur, Rastislav Bod{\'{\i}}k, Garvit Juniwal, Milo M.~K. Martin, Mukund
  Raghothaman, Sanjit~A. Seshia, Rishabh Singh, Armando Solar{-}Lezama, Emina
  Torlak, and Abhishek Udupa.
\newblock Syntax-guided synthesis.
\newblock In \emph{Formal Methods in Computer-Aided Design ({FMCAD})}, pp.\
  1--8, 2013.

\bibitem[Balog et~al.(2017)Balog, Gaunt, Brockschmidt, Nowozin, and
  Tarlow]{balog_iclr16}
Matej Balog, Alexander~L Gaunt, Marc Brockschmidt, Sebastian Nowozin, and
  Daniel Tarlow.
\newblock {DeepCoder}: Learning to write programs.
\newblock In \emph{International Conference on Learning Representations
  ({ICLR})}, 2017.

\bibitem[Bosnjak et~al.(2017)Bosnjak, Rockt{\"{a}}schel, Naradowsky, and
  Riedel]{riedel_arxiv16}
Matko Bosnjak, Tim Rockt{\"{a}}schel, Jason Naradowsky, and Sebastian Riedel.
\newblock Programming with a differentiable {Forth} interpreter.
\newblock In \emph{Proceedings of the 34th International Conference on Machine
  Learning, {ICML} 2017, Sydney, NSW, Australia, 6-11 August 2017}, pp.\
  547--556, 2017.

\bibitem[Cai et~al.(2017)Cai, Shin, and Song]{cai_iclr17}
Jonathon Cai, Richard Shin, and Dawn Song.
\newblock Making neural programming architectures generalize via recursion.
\newblock In \emph{International Conference on Learning Representations
  ({ICLR})}, 2017.

\bibitem[Chaudhari \& Damani(2015)Chaudhari and Damani]{chaudhari2015combining}
Dipak~L Chaudhari and Om~Damani.
\newblock Combining top-down and bottom-up techniques in program derivation.
\newblock In \emph{International Symposium on Logic-Based Program Synthesis and
  Transformation}, pp.\  244--258. Springer, 2015.

\bibitem[Clausen(1999)]{clausen_branchnbound}
Jens Clausen.
\newblock Branch and bound algorithms -- principles and examples.
\newblock \emph{Department of Computer Science, University of Copenhagen},
  1999.

\bibitem[Devlin et~al.(2017)Devlin, Uesato, Bhupatiraju, Singh, Mohamed, and
  Kohli]{devlin_icml17}
Jacob Devlin, Jonathan Uesato, Surya Bhupatiraju, Rishabh Singh, Abdel-rahman
  Mohamed, and Pushmeet Kohli.
\newblock {RobustFill}: Neural program learning under noisy {I/O}.
\newblock In \emph{International Conference on Machine Learning ({ICML})},
  2017.

\bibitem[Ellis \& Gulwani(2017)Ellis and Gulwani]{ellis2017learning}
Kevin Ellis and Sumit Gulwani.
\newblock Learning to learn programs from examples: Going beyond program
  structure.
\newblock In \emph{International Joint Conference on Artifical Intelligence
  ({IJCAI})}, 2017.

\bibitem[Gaunt et~al.(2016)Gaunt, Brockschmidt, Singh, Kushman, Kohli, Taylor,
  and Tarlow]{gaunt_arxiv16}
Alexander~L Gaunt, Marc Brockschmidt, Rishabh Singh, Nate Kushman, Pushmeet
  Kohli, Jonathan Taylor, and Daniel Tarlow.
\newblock {TerpreT}: A probabilistic programming language for program
  induction.
\newblock \emph{CoRR}, abs/1608.04428, 2016.
\newblock URL \url{http://arxiv.org/abs/1608.04428}.

\bibitem[Graves et~al.(2014)Graves, Wayne, and Danihelka]{graves_arxiv14}
Alex Graves, Greg Wayne, and Ivo Danihelka.
\newblock Neural {Turing} machines.
\newblock \emph{CoRR}, abs/1410.5401, 2014.
\newblock URL \url{http://arxiv.org/abs/1410.5401}.

\bibitem[Gulwani(2011)]{gulwani_popl11}
Sumit Gulwani.
\newblock Automating string processing in spreadsheets using input-output
  examples.
\newblock In \emph{Principles of Programming Languages ({POPL})}, volume~46,
  pp.\  317--330, 2011.

\bibitem[Gulwani \& Jain(2017)Gulwani and Jain]{gulwani2017programming}
Sumit Gulwani and Prateek Jain.
\newblock Programming by examples: Pl meets ml.
\newblock In \emph{Asian Symposium on Programming Languages and Systems}, pp.\
  3--20. Springer, 2017.

\bibitem[Gulwani et~al.(2017)Gulwani, Polozov, and Singh]{GulwaniPS17}
Sumit Gulwani, Oleksandr Polozov, and Rishabh Singh.
\newblock Program synthesis.
\newblock \emph{Foundations and Trends in Programming Languages}, 4\penalty0
  (1-2):\penalty0 1--119, 2017.
\newblock \doi{10.1561/2500000010}.
\newblock URL \url{https://doi.org/10.1561/2500000010}.

\bibitem[Hochreiter \& Schmidhuber(1997)Hochreiter and
  Schmidhuber]{hochreiter97}
Sepp Hochreiter and J\"{u}rgen Schmidhuber.
\newblock Long short-term memory.
\newblock \emph{Neural Comput.}, 9\penalty0 (8):\penalty0 1735--1780, November
  1997.
\newblock ISSN 0899-7667.
\newblock \doi{10.1162/neco.1997.9.8.1735}.
\newblock URL \url{http://dx.doi.org/10.1162/neco.1997.9.8.1735}.

\bibitem[Kaiser \& Sutskever(2015)Kaiser and Sutskever]{kaiser_arxiv15}
{\L}ukasz Kaiser and Ilya Sutskever.
\newblock Neural {GPUs} learn algorithms.
\newblock \emph{CoRR}, abs/1511.08228, 2015.
\newblock URL \url{http://arxiv.org/abs/1511.08228}.

\bibitem[Kingma \& Ba(2014)Kingma and Ba]{kingma2014adam}
Diederik Kingma and Jimmy Ba.
\newblock Adam: A method for stochastic optimization.
\newblock In \emph{International Conference on Learning Representations
  ({ICLR})}, 2014.

\bibitem[Le \& Gulwani(2014)Le and Gulwani]{le2014flashextract}
Vu~Le and Sumit Gulwani.
\newblock {FlashExtract}: A framework for data extraction by examples.
\newblock In \emph{ACM SIGPLAN Notices}, volume~49, pp.\  542--553. ACM, 2014.

\bibitem[Lin et~al.(2014)Lin, Dechter, Ellis, Tenenbaum, and
  Muggleton]{lin2014bias}
Dianhuan Lin, Eyal Dechter, Kevin Ellis, Joshua Tenenbaum, and Stephen
  Muggleton.
\newblock Bias reformulation for one-shot function induction.
\newblock In \emph{Proceedings of the Twenty-first European Conference on
  Artificial Intelligence}, pp.\  525--530. IOS Press, 2014.

\bibitem[Loos et~al.(2017)Loos, Irving, Szegedy, and Kaliszyk]{irving_arxiv16}
Sarah~M. Loos, Geoffrey Irving, Christian Szegedy, and Cezary Kaliszyk.
\newblock Deep network guided proof search.
\newblock In \emph{LPAR-21, 21st International Conference on Logic for
  Programming, Artificial Intelligence and Reasoning, Maun, Botswana, 7-12th
  May 2017}, pp.\  85--105, 2017.

\bibitem[Manna \& Waldinger(1971)Manna and Waldinger]{manna71}
Zohar Manna and Richard~J. Waldinger.
\newblock Toward automatic program synthesis.
\newblock \emph{Communications of the {ACM}}, 14\penalty0 (3):\penalty0
  151--165, 1971.

\bibitem[Parisotto et~al.(2016)Parisotto, Mohamed, Singh, Li, Zhou, and
  Kohli]{parisotto_iclr16}
Emilio Parisotto, Abdel-rahman Mohamed, Rishabh Singh, Lihong Li, Dengyong
  Zhou, and Pushmeet Kohli.
\newblock Neuro-symbolic program synthesis.
\newblock In \emph{International Conference on Learning Representations
  ({ICLR})}, 2016.

\bibitem[Polozov \& Gulwani(2015)Polozov and Gulwani]{polozov_oopsla15}
Oleksandr Polozov and Sumit Gulwani.
\newblock {FlashMeta}: A framework for inductive program synthesis.
\newblock In \emph{International Conference on Object-Oriented Programming,
  Systems, Languages, and Applications ({OOPSLA})}, pp.\  107--126, 2015.

\bibitem[Puschel et~al.(2005)Puschel, Moura, Johnson, Padua, Veloso, Singer,
  Xiong, Franchetti, Gacic, Voronenko, et~al.]{spiral}
Markus Puschel, Jos{\'e}~MF Moura, Jeremy~R Johnson, David Padua, Manuela~M
  Veloso, Bryan~W Singer, Jianxin Xiong, Franz Franchetti, Aca Gacic, Yevgen
  Voronenko, et~al.
\newblock {SPIRAL}: Code generation for {DSP} transforms.
\newblock \emph{Proceedings of the IEEE}, 93\penalty0 (2):\penalty0 232--275,
  2005.

\bibitem[Reed \& De~Freitas(2016)Reed and De~Freitas]{reed_arxiv15}
Scott Reed and Nando De~Freitas.
\newblock Neural programmer-interpreters.
\newblock In \emph{International Conference on Learning Representations
  ({ICLR})}, 2016.

\bibitem[Rolim et~al.(2017)Rolim, Soares, D'Antoni, Polozov, Gulwani, Gheyi,
  Suzuki, and Hartmann]{rolim2017learning}
Reudismam Rolim, Gustavo Soares, Loris D'Antoni, Oleksandr Polozov, Sumit
  Gulwani, Rohit Gheyi, Ryo Suzuki, and Bj{\"o}rn Hartmann.
\newblock Learning syntactic program transformations from examples.
\newblock In \emph{International Conference on Software Engineering ({ICSE})},
  pp.\  404--415, 2017.

\bibitem[Seide \& Agarwal(2016)Seide and Agarwal]{cntk}
Frank Seide and Amit Agarwal.
\newblock {CNTK}: Microsoft's open-source deep-learning toolkit.
\newblock In \emph{International Conference on Knowledge Discovery and Data
  Mining ({KDD})}, pp.\  2135--2135, 2016.

\bibitem[Singh \& Gulwani(2015)Singh and Gulwani]{singh_cav15}
Rishabh Singh and Sumit Gulwani.
\newblock Predicting a correct program in programming by example.
\newblock In \emph{Computer-Aided Verification ({CAV})}, 2015.

\bibitem[Torlak \& Bodik(2013)Torlak and Bodik]{torlak2013growing}
Emina Torlak and Rastislav Bodik.
\newblock Growing solver-aided languages with {Rosette}.
\newblock In \emph{Proceedings of the 2013 ACM international symposium on New
  ideas, new paradigms, and reflections on programming \& software}, pp.\
  135--152. ACM, 2013.

\bibitem[Udupa et~al.(2013)Udupa, Raghavan, Deshmukh, Mador-Haim, Martin, and
  Alur]{transit}
Abhishek Udupa, Arun Raghavan, Jyotirmoy~V. Deshmukh, Sela Mador-Haim,
  Milo~M.K. Martin, and Rajeev Alur.
\newblock {TRANSIT}: Specifying protocols with concolic snippets.
\newblock In \emph{Programming Languages Design and Implementation (PLDI)},
  pp.\  287--296, 2013.

\bibitem[Waldinger \& Lee(1969)Waldinger and Lee]{waldinger1969prow}
Richard~J Waldinger and Richard~CT Lee.
\newblock {PROW}: A step toward automatic program writing.
\newblock In \emph{International Joint Conference on Artificial Intelligence
  ({IJCAI})}, pp.\  241--252, 1969.

\bibitem[Zaremba et~al.(2016)Zaremba, Mikolov, Joulin, and
  Fergus]{zaremba_icml16}
Wojciech Zaremba, Tomas Mikolov, Armand Joulin, and Rob Fergus.
\newblock Learning simple algorithms from examples.
\newblock In \emph{International Conference on Machine Learning ({ICML})},
  2016.

\end{thebibliography}
\bibliographystyle{iclr2018_conference}
\clearpage

\appendix
\section{RobustFill Performance with Different Beam Sizes}\label{app:rfill}
For our experiments, we implemented RobustFill with the beam size of 100, as it presented a good trade-off between
generalization accuracy and performance hit.
The following table shows a detailed comparison of RobustFill's generalization accuracy and performance for different
beam sizes and numbers of training examples.

\begin{table}[h]
    \centering
    \begin{tabular}{llrr}
        \toprule
        \bfseries Number of examples ($\bm{m}$) & \bfseries Beam size & \bfseries Accuracy ($\pmb{\%}$) & \bfseries Speed-up ($\bm{\times}$ PROSE)\\
        \midrule
        \multirow{3}{*}{\textbf{1}} &
        10&18.4&0.41\\
          &100&24.5&0.25\\
          &1000&34.1&0.04\\
        \midrule
        \multirow{3}{*}{\textbf{2}} &
        10&32.2&0.43\\
          &100&39.7&0.27\\
          &1000&47.6&0.04\\
        \midrule
        \multirow{3}{*}{\textbf{3}} &
        10&49.8&0.48\\
          &100&56.4&0.30\\
          &1000&63.4&0.04\\
        \bottomrule
    \end{tabular}
\caption{Generalization accuracy and performance of RobustFill for different
	beam sizes and numbers of training examples.}
\label{table:robustfill}
\end{table}

\section{Performance of Best NGDS Model on All Non-Training Tasks}
\label{sec:benchmark-table}
\newcommand{\correct}{\textcolor{Green}{\cmark}}
\newcommand{\wrong}{\textcolor{Red}{\xmark}}
{\small
\setlength\tabcolsep{4pt}
\begin{longtable}[c]{@{}llrrrcc@{}}
    \toprule
    \textbf{Task \#} & \textbf{Test/Val}  & \textbf{PROSE Time (s)}     & \textbf{NGDS Time (s)}      & \textbf{Speed-up}     & \textbf{PROSE
    Correct?} & \textbf{NGDS Correct?} \\* \midrule
    \endhead
    \bottomrule
    \endfoot
    \endlastfoot
    1 & Test & 3.0032564 & 0.233686 & 12.85167 & \correct & \correct \\
    2 & Validation & 1.1687841 & 0.211069 & 5.53745 & \correct & \wrong \\
    3 & Validation & 0.4490832 & 0.1307367 & 3.43502 & \correct & \correct \\
    4 & Test & 6.665234 & 2.012157 & 3.312482 & \correct & \wrong \\
    5 & Test & 2.28298 & 0.83715 & 2.727086 & \wrong & \wrong \\
    6 & Test & 3.0391034 & 1.1410092 & 2.663522 & \correct & \wrong \\
    7 & Validation & 0.5487662 & 0.2105728 & 2.606064 & \correct & \correct \\
    8 & Test & 2.4120103 & 0.9588959 & 2.515404 & \wrong & \wrong \\
    9 & Validation & 7.6010733 & 3.052303 & 2.490275 & \wrong & \wrong \\
    10 & Test & 2.1165486 & 0.8816776 & 2.400592 & \wrong & \wrong \\
    11 & Test & 0.9622929 & 0.405093 & 2.375486 & \correct & \correct \\
    12 & Validation & 0.4033455 & 0.1936532 & 2.082824 & \wrong & \wrong \\
    13 & Test & 0.4012993 & 0.1929299 & 2.080026 & \correct & \correct \\
    14 & Validation & 2.9467418 & 1.4314372 & 2.05859 & \correct & \correct \\
    15 & Test & 0.3855433 & 0.1987497 & 1.939843 & \wrong & \wrong \\
    16 & Test & 6.0043011 & 3.1862577 & 1.884437 & \wrong & \wrong \\
    17 & Test & 3.0316721 & 1.6633142 & 1.82267 & \wrong & \wrong \\
    18 & Test & 0.3414629 & 0.1933263 & 1.766252 & \correct & \correct \\
    19 & Validation & 0.3454594 & 0.2014236 & 1.715089 & \correct & \correct \\
    20 & Test & 0.3185586 & 0.202928 & 1.569811 & \wrong & \wrong \\
    21 & Test & 0.2709963 & 0.1734634 & 1.562268 & \correct & \correct \\
    22 & Test & 0.4859534 & 0.3169533 & 1.533202 & \correct & \correct \\
    23 & Test & 0.8672071 & 0.5865048 & 1.478602 & \correct & \wrong \\
    24 & Validation & 0.3626161 & 0.2590434 & 1.399828 & \correct & \correct \\
    25 & Validation & 2.3343791 & 1.6800684 & 1.389455 & \correct & \correct \\
    26 & Test & 0.2310051 & 0.1718745 & 1.344034 & \correct & \correct \\
    27 & Test & 0.1950921 & 0.1456817 & 1.339167 & \correct & \correct \\
    28 & Test & 0.8475303 & 0.6425532 & 1.319004 & \correct & \correct \\
    29 & Validation & 0.4064375 & 0.316499 & 1.284167 & \correct & \correct \\
    30 & Test & 0.2601689 & 0.2083826 & 1.248515 & \wrong & \wrong \\
    31 & Test & 0.2097732 & 0.1753706 & 1.196171 & \correct & \correct \\
    32 & Test & 1.2224533 & 1.0264273 & 1.190979 & \wrong & \wrong \\
    33 & Test & 0.5431827 & 0.4691296 & 1.157852 & \correct & \correct \\
    34 & Validation & 0.4183223 & 0.3685321 & 1.135104 & \correct & \correct \\
    35 & Validation & 0.2497723 & 0.2214195 & 1.12805 & \wrong & \correct \\
    36 & Validation & 0.2385918 & 0.212407 & 1.123277 & \wrong & \wrong \\
    37 & Test & 0.2241414 & 0.2004937 & 1.117947 & \correct & \correct \\
    38 & Validation & 0.2079995 & 0.1880859 & 1.105875 & \correct & \correct \\
    39 & Test & 0.2788713 & 0.2654384 & 1.050606 & \correct & \correct \\
    40 & Test & 0.1821743 & 0.1758255 & 1.036109 & \correct & \correct \\
    41 & Validation & 0.1486939 & 0.1456755 & 1.02072 & \correct & \correct \\
    42 & Test & 0.3981185 & 0.3900767 & 1.020616 & \wrong & \correct \\
    43 & Test & 0.9959218 & 0.9960901 & 0.999831 & \correct & \correct \\
    44 & Test & 0.2174055 & 0.2239088 & 0.970956 & \correct & \correct \\
    45 & Test & 1.8684116 & 1.9473475 & 0.959465 & \correct & \correct \\
    46 & Test & 0.1357812 & 0.1428591 & 0.950455 & \correct & \correct \\
    47 & Validation & 0.2549691 & 0.2709866 & 0.940892 & \wrong & \wrong \\
    48 & Test & 0.1650636 & 0.1762617 & 0.936469 & \correct & \correct \\
    49 & Validation & 0.5368683 & 0.5781537 & 0.928591 & \correct & \correct \\
    50 & Test & 0.1640937 & 0.1851361 & 0.886341 & \wrong & \wrong \\
    51 & Validation & 0.5006552 & 0.5736976 & 0.872681 & \correct & \correct \\
    52 & Test & 0.2064185 & 0.2401594 & 0.859506 & \correct & \correct \\
    53 & Validation & 0.2381335 & 0.277788 & 0.857249 & \wrong & \wrong \\
    54 & Test & 0.2171637 & 0.2677121 & 0.811184 & \correct & \correct \\
    55 & Test & 0.6307356 & 0.7807711 & 0.807837 & \correct & \correct \\
    56 & Validation & 0.3462029 & 0.4325302 & 0.800413 & \correct & \correct \\
    57 & Test & 0.4285604 & 0.5464594 & 0.784249 & \wrong & \wrong \\
    58 & Validation & 0.155915 & 0.1992245 & 0.78261 & \correct & \correct \\
    59 & Test & 0.1651815 & 0.2135129 & 0.773637 & \wrong & \wrong \\
    60 & Validation & 0.1212689 & 0.1571558 & 0.771648 & \correct & \correct \\
    61 & Test & 0.1980844 & 0.257616 & 0.768913 & \correct & \correct \\
    62 & Validation & 0.1534717 & 0.2004651 & 0.765578 & \correct & \correct \\
    63 & Test & 0.2443636 & 0.3258476 & 0.749932 & \wrong & \wrong \\
    64 & Test & 0.1217696 & 0.1635984 & 0.74432 & \correct & \correct \\
    65 & Validation & 0.2446501 & 0.3301224 & 0.741089 & \correct & \correct \\
    66 & Validation & 0.6579789 & 0.8886647 & 0.740413 & \correct & \wrong \\
    67 & Test & 0.1490806 & 0.2022204 & 0.737218 & \correct & \correct \\
    68 & Test & 0.2668753 & 0.3681659 & 0.724878 & \correct & \correct \\
    69 & Test & 0.1072814 & 0.1487589 & 0.721176 & \correct & \correct \\
    70 & Validation & 0.1310034 & 0.181912 & 0.720147 & \correct & \wrong \\
    71 & Test & 0.1954476 & 0.273414 & 0.714841 & \correct & \correct \\
    72 & Test & 0.3323319 & 0.468445 & 0.709436 & \correct & \correct \\
    73 & Test & 0.2679471 & 0.3806013 & 0.70401 & \correct & \correct \\
    74 & Test & 1.1505939 & 1.6429378 & 0.700327 & \correct & \correct \\
    75 & Test & 0.1318375 & 0.1898685 & 0.694362 & \wrong & \wrong \\
    76 & Test & 0.15018 & 0.2189491 & 0.685913 & \wrong & \wrong \\
    77 & Test & 0.146774 & 0.2144594 & 0.684391 & \correct & \correct \\
    78 & Test & 0.1123303 & 0.1665129 & 0.674604 & \correct & \correct \\
    79 & Test & 0.1623439 & 0.2468262 & 0.657726 & \wrong & \wrong \\
    80 & Test & 0.4243661 & 0.6563517 & 0.646553 & \wrong & \wrong \\
    81 & Test & 0.2945639 & 0.4662018 & 0.631838 & \wrong & \correct \\
    82 & Validation & 0.0892761 & 0.1419142 & 0.629085 & \correct & \wrong \\
    83 & Test & 0.1992316 & 0.3229269 & 0.616956 & \correct & \correct \\
    84 & Validation & 0.3260828 & 0.5294719 & 0.615864 & \correct & \correct \\
    85 & Test & 0.2181703 & 0.3576818 & 0.609956 & \correct & \correct \\
    86 & Test & 0.1757585 & 0.3006565 & 0.584582 & \correct & \correct \\
    87 & Validation & 0.1811467 & 0.3107196 & 0.582991 & \correct & \correct \\
    88 & Test & 0.2774191 & 0.4759698 & 0.58285 & \wrong & \correct \\
    89 & Test & 0.137414 & 0.2358583 & 0.582613 & \correct & \correct \\
    90 & Validation & 0.1051238 & 0.1834589 & 0.57301 & \correct & \correct \\
    91 & Validation & 1.5624891 & 2.7446374 & 0.569288 & \correct & \correct \\
    92 & Validation & 0.1104184 & 0.1958337 & 0.563838 & \wrong & \wrong \\
    93 & Validation & 0.1233551 & 0.2228252 & 0.553596 & \wrong & \wrong \\
    94 & Validation & 0.189019 & 0.3445496 & 0.548597 & \wrong & \wrong \\
    95 & Validation & 0.2997031 & 0.5486731 & 0.546233 & \correct & \correct \\
    96 & Test & 0.1057559 & 0.19453 & 0.543648 & \correct & \correct \\
    97 & Validation & 0.129731 & 0.2426926 & 0.534549 & \correct & \correct \\
    98 & Test & 0.1706376 & 0.320323 & 0.532705 & \correct & \correct \\
    99 & Test & 0.0936175 & 0.1764753 & 0.530485 & \correct & \correct \\
    100 & Test & 0.2101397 & 0.40277 & 0.521736 & \wrong & \wrong \\
    101 & Test & 0.1816704 & 0.3507656 & 0.517925 & \correct & \correct \\
    102 & Validation & 0.1516109 & 0.2993282 & 0.506504 & \correct & \correct \\
    103 & Test & 0.1102942 & 0.2185006 & 0.504778 & \correct & \correct \\
    104 & Validation & 1.1538661 & 2.3299578 & 0.49523 & \wrong & \correct \\
    105 & Test & 0.1241092 & 0.251046 & 0.494368 & \wrong & \correct \\
    106 & Test & 1.068263 & 2.176145 & 0.490897 & \wrong & \wrong \\
    107 & Validation & 0.1899474 & 0.389012 & 0.488282 & \correct & \correct \\
    108 & Validation & 0.205652 & 0.4312716 & 0.47685 & \correct & \wrong \\
    109 & Test & 0.1332348 & 0.2819654 & 0.472522 & \correct & \correct \\
    110 & Test & 0.2137989 & 0.4625152 & 0.462253 & \wrong & \wrong \\
    111 & Validation & 0.2233911 & 0.4898705 & 0.456021 & \wrong & \wrong \\
    112 & Validation & 0.1742123 & 0.3872159 & 0.44991 & \correct & \correct \\
    113 & Test & 0.1798306 & 0.4059525 & 0.442984 & \correct & \correct \\
    114 & Validation & 0.1576141 & 0.3592128 & 0.438776 & \correct & \correct \\
    115 & Test & 0.1441545 & 0.3462711 & 0.416305 & \correct & \correct \\
    116 & Validation & 0.189833 & 0.4649153 & 0.408317 & \wrong & \wrong \\
    117 & Validation & 0.3401477 & 1.0468088 & 0.324938 & \correct & \correct \\
    118 & Validation & 0.1575744 & 0.6015111 & 0.261964 & \wrong & \wrong \\
    119 & Validation & 0.7252624 & 3.2088775 & 0.226017 & \correct & \wrong \\
    120 & Test & 0.1288099 & 0.5958986 & 0.216161 & \correct & \correct \\* \bottomrule
\end{longtable}
}

\section{ML-based Ranker}
As noted in \cref{sec:background}, learning a ranking function is an interesting problem in itself and is orthogonal to our work.
Since our method can be used along with any accurate ranking function, we assume black-box access to such a high-quality ranker and specifically, use the state-of-the-art ranking function of PROSE that involves a significant amount of hand engineering.
\\ \\
In this section, we evaluate the performance of our method and PROSE when employing a competitive ranker learned in a data-driven manner \citep{gulwani2017programming}.
From the table below, it can be observed that when using an ML-based ranking function, our method achieves an average
$\approx 2\times$ speed-up over PROSE while still achieving comparable generalization accuracy .
\begin{table}[h]
    \centering
    \begin{tabular}{lrrr}
        \toprule
        \textbf{Metric} &\textbf{PROSE} & \textbf{NGDS ($\bm{T_1}$, BB)} & \textbf{NGDS ($\bm{T_1 + POS}$, BB)} \\
        \midrule
        \textbf{Accuracy (\% of 73)} & \textbf{65.75} & \textbf{65.75} & 64.38 \\
        \textbf{Speed-up ($\times$ PROSE)}  & 1.00 & 2.15 & \textbf{2.46} \\
        \bottomrule
    \end{tabular}
    \caption{Generalization accuracy and speed-up of NGDS variants vs. PROSE where all methods use a machine learning based ranking function from \cite{gulwani2017programming}.}
\end{table}

\end{document}